\documentclass[3p,times]{elsarticle}

\usepackage{ecrc}
\usepackage{caption}
\usepackage{subcaption}
\usepackage{amsmath}

\volume{00}

\firstpage{1}

\journalname{Remote Sensing of Environment}

\runauth{}

\jid{RSE}

\CopyrightLine{2011}{Published by Elsevier Ltd.}

\usepackage{amssymb}

\usepackage[figuresright]{rotating}

\begin{document}

\begin{frontmatter}



\dochead{Short communication}

\title{A Recommender System-Inspired Cloud Data Filling Scheme for Satellite-based Coastal Observation}


\author{Ruo-Qian Wang\corref{cor1}}
\ead{rq.wang@rutgers.edu}
\cortext[cor1]{}
\address{Department of Civil and Environmental Engineering, Rutgers, the State University of New Jersey, New Jersey, USA 08816}

\begin{abstract}
Filling missing data in cloud-covered areas of satellite imaging is an important task to improve data quantity and quality for enhanced earth observation. Traditional cloud filling studies focused on continuous numerical data such as temperature and cyanobacterial concentration in the open ocean. Cloud data filling issues in coastal imaging is far less studied because of the complex landscape. Inspired by the success of data imputation methods in recommender systems that are designed for online shopping, the present study explored their application to satellite cloud data filling tasks. A numerical experiment was designed and conducted for a LandSat dataset with a range of synthetic cloud covers to examine the performance of different data filling schemes. The recommender system-inspired matrix factorization algorithm called Funk-SVD showed superior performance in computational accuracy and efficiency for the task of recovering landscape types in a complex coastal area than the traditional data filling scheme of DINEOF (Data Interpolating Empirical Orthogonal Functions) and the deep learning method of Datawig. The new method achieved the best filling accuracy and reached a speed comparable to DINEOF and much faster than deep learning. A theoretical framework was created to analyze the error propagation in DINEOF and found the algorithm needs to be modified to converge to the ground truth. The present study showed that Funk-SVD has great potential to enhance cloud data filling performance and connects the fields of recommender systems and cloud filling to promote the improvement and sharing of useful algorithms.
\end{abstract}


\end{frontmatter}

\vspace{0.5cm}
After decades of development, satellite-based sensing technologies now provide exciting opportunities to make continuous and wide observation of the earth's surface. A remarkable example is the series of LandSat missions, which captures the imagery of global land conditions and dynamics every 14 days since July 1972 \cite{loveland_landsat_2012}. However, the application of the obtained data is limited, especially in the coastal area where significant cloud cover presents. The cloud cover reduces the spatial-temporal availability of earth observation, causes the problem of data gap and noise, and poses challenges in earth surface analysis.

Several data processing schemes have been developed to fill cloud data gaps and they are demonstrated, in general, helpful to improve earth observation. For example, a recent study found that applying cloud filling schemes can improve the accuracy of the mean algal concentration estimate by 50\% to 80\% \cite{stock_comparison_2020}. The principle of cloud data filling algorithms is to predict the earth surface dynamics blocked by cloud covers in remote sensing using the cloud-free data available at other times or locations, e.g., the data of the cloud-covered area obtained at the cloud-free time or the data of the neighboring cloud-free locations at the time of the cloud cover. The most widely used scheme is probably DINEOF (Data-Interpolating Empirical Orthogonal Functions) \cite{beckers_dineof_2006}. Based on the feature of Empirical Orthogonal Functions that the leading modes of the observation data represent low-frequency large-scale structures, DINEOF truncates the lower-ranked, local and high-frequency modes to interpolate the missing data. This method has been widely applied to fill cloud data in various fields \cite{sirjacobs_cloud_2011,beckers_dineof_2006, alvera-azcarate_analysis_2015,alvera-azcarate_reconstruction_2005,alvera-azcarate_correction_2007,li_spatial_2014, liu_filling_2019, hilborn_applications_2018}. Geostatistical interpolation is another common approach to fill cloud data gaps \cite{urquhart_geospatial_2013}, which leverages the statistical features of the available part of the dataset to estimate the missing values. For example, Saulquin et al. \cite{saulquin_interpolated_2019} developed a Kriging algorithm to provide a gap-free global Chlorophyll a data product for Copernicus, the European Union’s earth observation program. Recently, machine learning methods, e.g., self-organizing maps (SOMs) \cite{jouini_reconstruction_2013} and random forests \cite{chen_improving_2019, park_reconstruction_2019}, are becoming popular in data filling applications and shown to improve the data filling accuracy. A comparison among the popular methods for Chlorophyll a concentration in the open ocean showed that the machine learning methods are superior than the statistical methods \cite{stock_comparison_2020}.

However, the present cloud data filling schemes are still unsatisfactory in applications. First, the computational load of the machine learning-based algorithms is still heavy -- the intensive training could consume significant computational resources. Second, the fast-algorithms such as the matrix decomposition algorithms often have low accuracy, which limits their application in the area of complex landscapes. Third, most existing studies have focused on the application for the open ocean and continuous numerical variables, while limited experience has been gained about the issue of cloud cover in coastal areas where land and ocean meet to form complex landscapes involving categorical data such as the landscape type.

\begin{figure}
\centering
\includegraphics[width=0.6\linewidth]{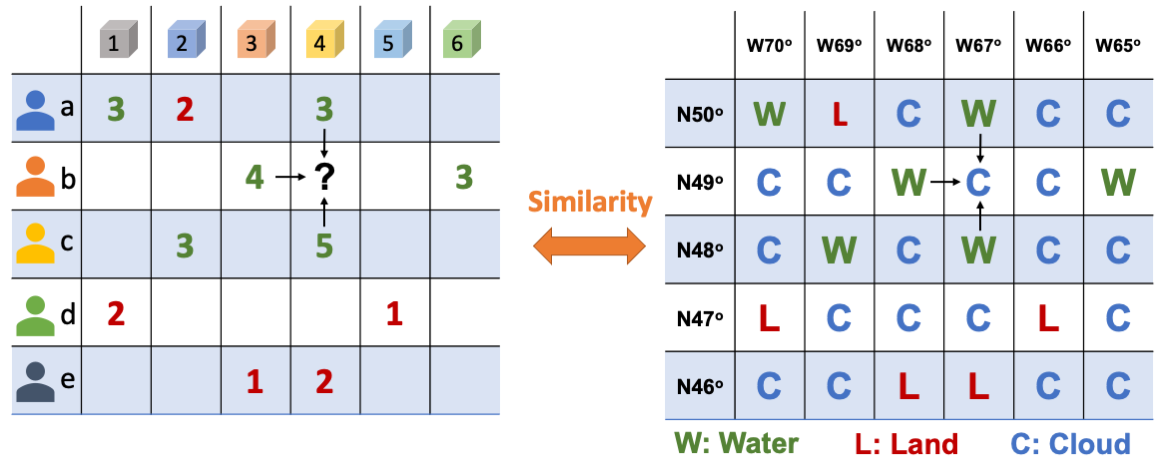}
\caption{The similarity is discovered between the recommender system and cloud filling schemes -- the element of the matrix in the recommender system (left) represents the recorded or potential rating of a product by a user, while the element in the matrix of cloud filling task (right) represents the landscape type of a pixel.}
\label{fig:similarity}
\end{figure}

The present study proposes a new cloud filling scheme inspired by the success of data imputation methods in recommender systems. Recommender systems and the algorithms behind them have stimulated the rise of e-service platforms such as Alibaba, Flipkart, Netflix, etc. Leveraging the large number of customers' records, these platforms are able to provide accurate and customized recommendation of new products or services to their customers to aid the decision making \cite{sivapalan_recommender_2014}: similar users are assumed to share the preference for similar products or services. Following this argument, data imputation is used to predict how much a user likes a potential product or service even though the user has never tried it. For example, in Figure \ref{fig:similarity}, User b has never purchased Product 4 and hence there has been no data about how much User b would rate Product 4. Using data imputation schemes, the platform could estimate the unknown rating using other similar users' ratings for Product 4 or User b's past ratings for other products similar to Product 4. This process essentially fills the data gap in the matrix with information available from the rest of the matrix. We recognize that the task is similar to cloud filling -- if a pixel is covered by cloud, it becomes a data gap and one could fill the missing element by interpolating using the rest of the matrix. Recognizing the similarity between the two tasks (Figure \ref{fig:similarity}), the proved algorithm in recommender systems can be extended to solve the cloud filling problem. In this study, a proven recommender system algorithm is explored in the task of filling cloud data gaps and compared with other cloud filling algorithms to examine its performance.


\section*{Training and Testing Data}
\begin{figure}
     \centering
     \begin{subfigure} {\linewidth}
         \centering
         \includegraphics[width=\textwidth]{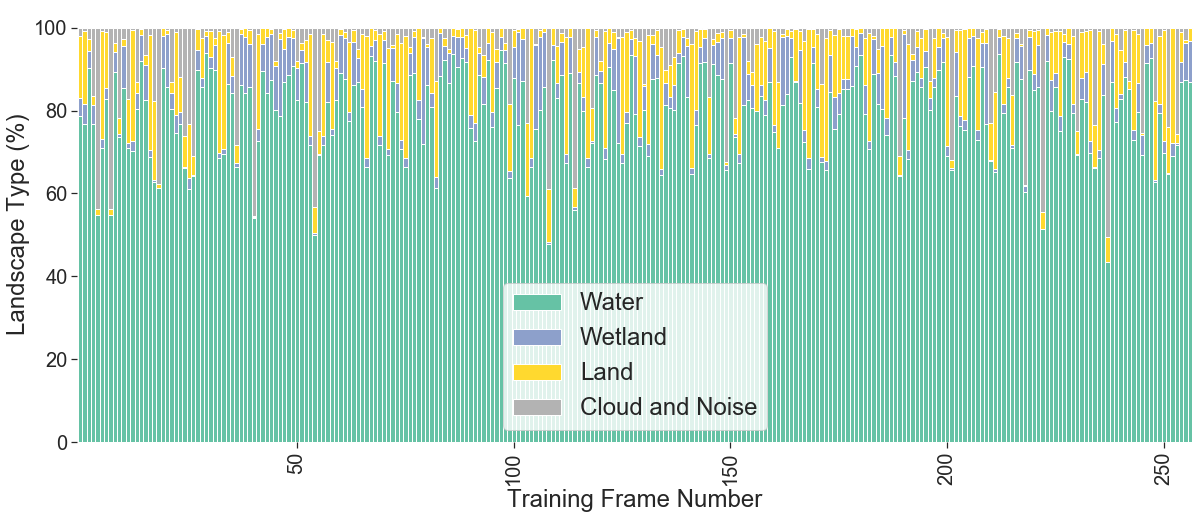}
         \caption*{(a) The percentage of landscape types for the 258 training frames.}
         \label{fig:y equals x}
     \end{subfigure}
     \hspace{10pts}
     \begin{subfigure} {\linewidth}
         \centering
         \includegraphics[width=\textwidth]{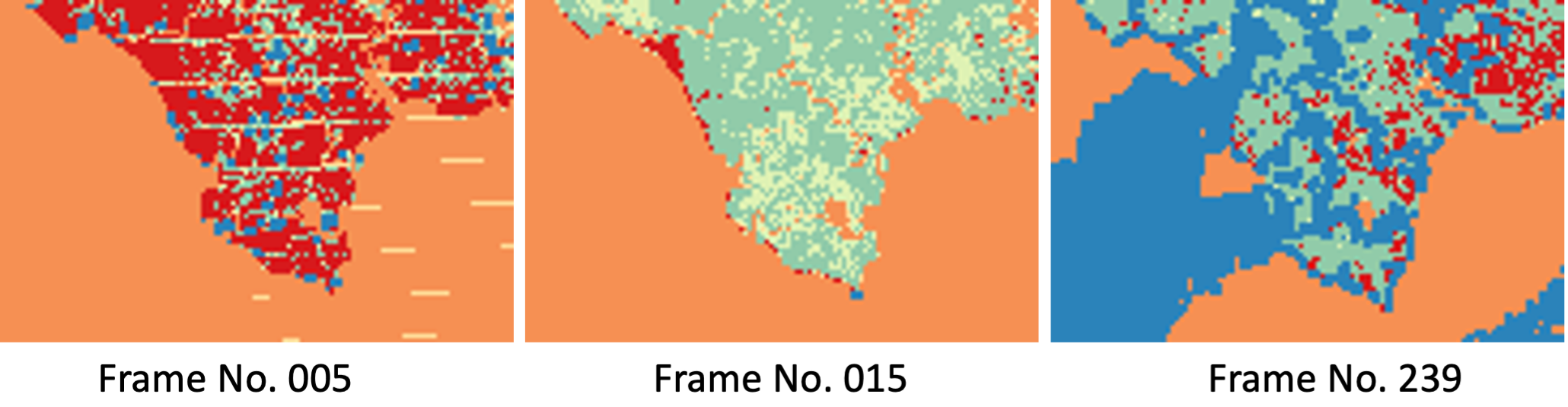}
         \caption*{(b) Sample data picked from the training data.}
         \label{fig:three sin x}
     \end{subfigure}
        \caption{The statistics and sample data from the training dataset.}
        \label{fig:training}
\end{figure}

An image database of 258 frames derived from LandSat missions for a small area in the Delaware Bay is selected to train and test the proposed cloud filling scheme. The landscape was classified into a variety of types following the United States Geological Survey (USGS)'s product -- Dynamic Water Surface Extent (DWSE) \cite{jones_improved_2019}. The percentage of land, water, wetland, and cloud and noise pixels in each frame is shown in Figure \ref{fig:training}a. It shows that the cloud and noise percentage is up to 50.4\% and the average water percentage is around 70\%. The area of land and wetland could significantly vary between frames. Three sample snapshots are shown in Figure \ref{fig:training}b and they reflect the high complexity of the training data. Large temporal variation of the landscape is observed in the training data, which is contributed by the seasonal and other hydrological/oceanographic factors. Simple interpolation schemes could not fill the missing data for such a challenging dataset with high nonlinearity.

\begin{figure}
\centering
\includegraphics[width=0.9\linewidth]{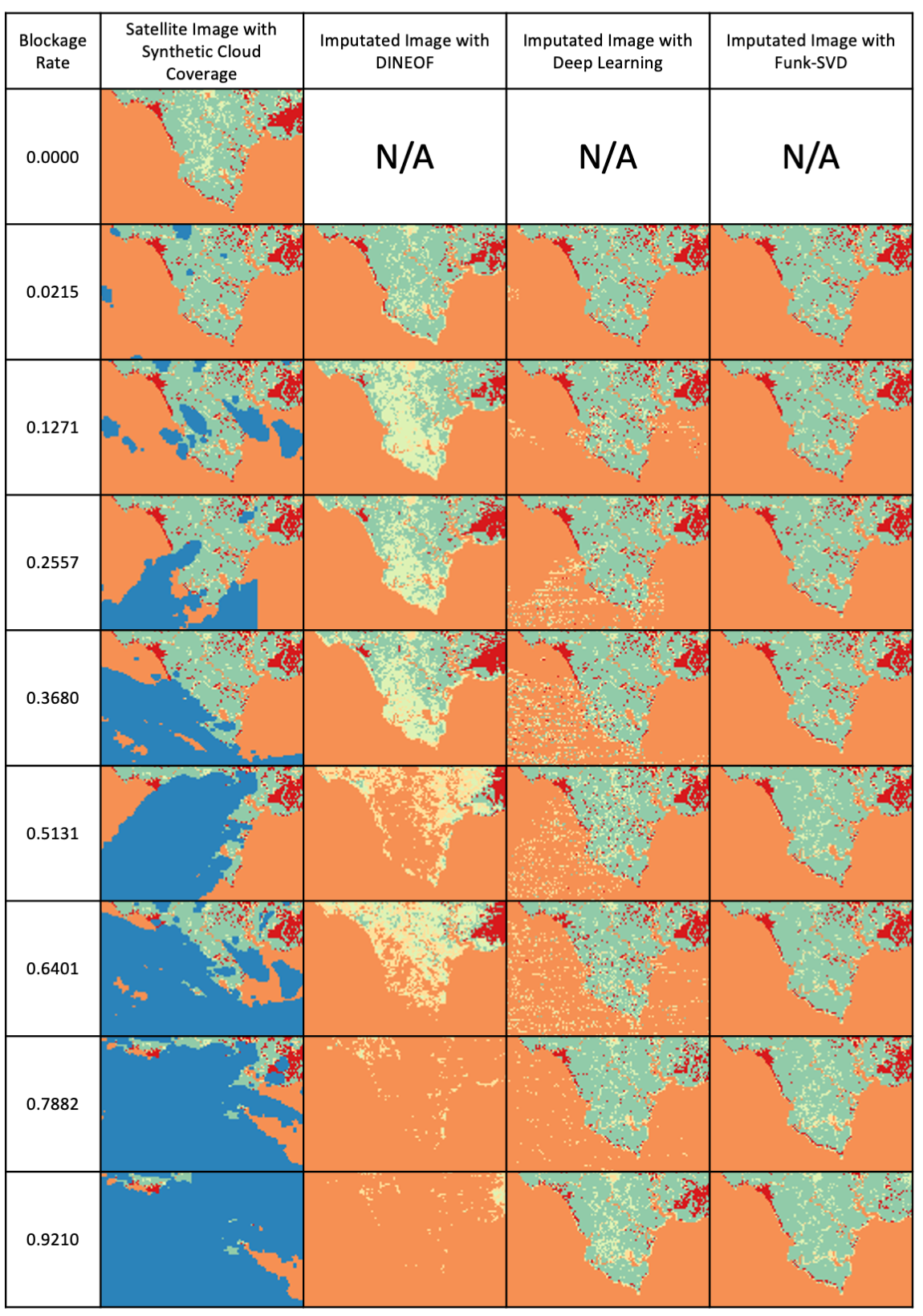}
\caption{The comparison of the cloud filling results among the different schemes.}
\label{fig:result}
\end{figure}

A frame of the landscape imaging (Figure \ref{fig:result}) originally free of cloud and not included in the training data is used as the ground truth to create synthetic cloud data for testing. Synthetic clouds copied from other time or locations are pasted to the ground truth frame so that random, but realistic-looking, synthetic clouds were created as the cloud covers. These examples with a wide range of cloud-cover rates (blocking rates) are used to test the performance of various schemes (the first column of Figure \ref{fig:result}). The training and ground truth data are available online \cite{wang_roger_dataset_2021}.

\section*{Methods}
To perform the cloud filling task, we arrange the imaging data into the matrix form: a series of $M$ satellite images, each with the size of $m\times n$, is flattened to a vector and they are stacked to form a matrix $\boldsymbol{X}_{M\times N}$ so that each row represents an image ($N=m\times n$). Every element of the matrix represents the landscape type of the location, and thus, forms an orderly categorical dataset. Integers are used to reflect the order of the data -- water, wetland, and land are labeled as 1, 2, and 3. The pixels covered by clouds are labeled as NaN (Not-a-number) to distinguish them from the available data.

Three popular cloud filling schemes were tested and compared in the present study. The first scheme is DINEOF, which is an EOF-based cloud data filling scheme and has been widely applied in analyzing ocean data. In this method, a matrix, $K_{M\times N}$, is created to capture the location of the missing data -- the cloud cover pixels are denoted $1$s and the others are denoted zeros, i.e.
\begin{equation}\label{eq:K}
    K_{M,N}=\left\{\begin{matrix}
 & 1  & for & x_{i,j,0}=NaN \\ 
 & 0  & for & x_{i,j,0}\neq NaN.
\end{matrix}\right.
\end{equation}

To start the operation, all the missing elements in the matrix are replaced by zeros and we name this new matrix $\mathbf{X}_{M\times N,0}$. An iteration scheme is applied following the steps below:

1) The matrix is decomposed by SVD
\begin{equation}
U_{M\times r,p}\Sigma_{r\times r,p}V_{N\times r,p}^T=\mathbf{X}_{M\times N,p},
\end{equation}
where $U$ are the spatial modes, $\Sigma$ is a diagonal matrix containing the singular values, and $V$ are the spatial modes.

2) Truncate the matrix into the largest $k$ ranks
\begin{equation}
\overline{\mathbf{X}}_{M\times N,p}^k=U_{M\times k,p}\Sigma_{k\times k,p}V_{N\times k,p}^T.
\end{equation}



3) Update the missing elements in the original matrix by the new elements at the missing places extracted from the truncated matrix,
\begin{equation}\label{eq:x_bar}
\mathbf{X}_{M\times N,p+1}=\mathbf{X}_{M\times N,0}+K_{M\times N}\overline{\mathbf{X}}_{M\times N, p}K_{M\times N}
\end{equation}

The iteration is repeated until the convergence is reached, i.e.
\begin{equation}
    \left\|\mathbf{X}_{M\times N,p+1}-\mathbf{X}_{M\times N,p}\right\|<\sigma,
\end{equation}
where $\sigma$ is a small number.

The second scheme is the emerging machine-learning based scheme. We selected a popular deep learning model called Datawig, which has been proved to be highly effective in data imputation. In this scheme, a column, $X_J$, that contains a missing data is first spotted. Then, the columns that have complete data, $\widetilde{X_J}$, is used to train the deep network model $f(X)$,
\begin{equation}
    \widehat{y}=f(\overline{X_J}),
\end{equation}
where $\overline{X_J}$ contains the rows of available data in Column $X_J$ and $\widehat{y}$ is used to compare with $\overline{X_J}$ to examine the training error. The trained function is next used to predict the missing values. This procedure is repeated until all the missing values are imputated \cite{biessmann_datawig_2019}.

The third scheme is the new, recommender system-inspired method called Funk-SVD published by Simon Funk on his blog \cite{funk_netflix_2006}. This scheme is designed to approximate the data matrix by the product of two lower rank matrices \cite{kumar_novel_2016, jiang_application_2020}, i.e.
\begin{equation}
    \mathbf{X}_{M\times N, 0}=\mathbf{P}_{M\times k}\times\mathbf{Q}_{k\times N}.
\end{equation}
A minimization scheme is employed to approximate every available element in the matrix of $\mathbf{X}_{M\times N, 0}$ while forcing the magnitude of the latent vectors $p_i$ and $q_j$ to be minimum,

\begin{equation}
    \min_{p,q}\sum_{i,j}(x_{i,j,0}-q^T_jp_i)^2+\lambda(\left \| p_i \right \|^2+\left \| q_j \right \|^2),
\end{equation}
where $p_i$ and $q_j$ are the columns of $\mathbf{P}_{M\times k}$ and $\mathbf{Q}_{k\times N}$ and $\lambda$ is the hyperparameter to tune the relative weight of the regularization.

The accuracy of the imputation is defined as the F1 score of the imputated image comparing to the ground truth.

\section*{Results}
\begin{figure}
\centering
\includegraphics[width=0.5\linewidth]{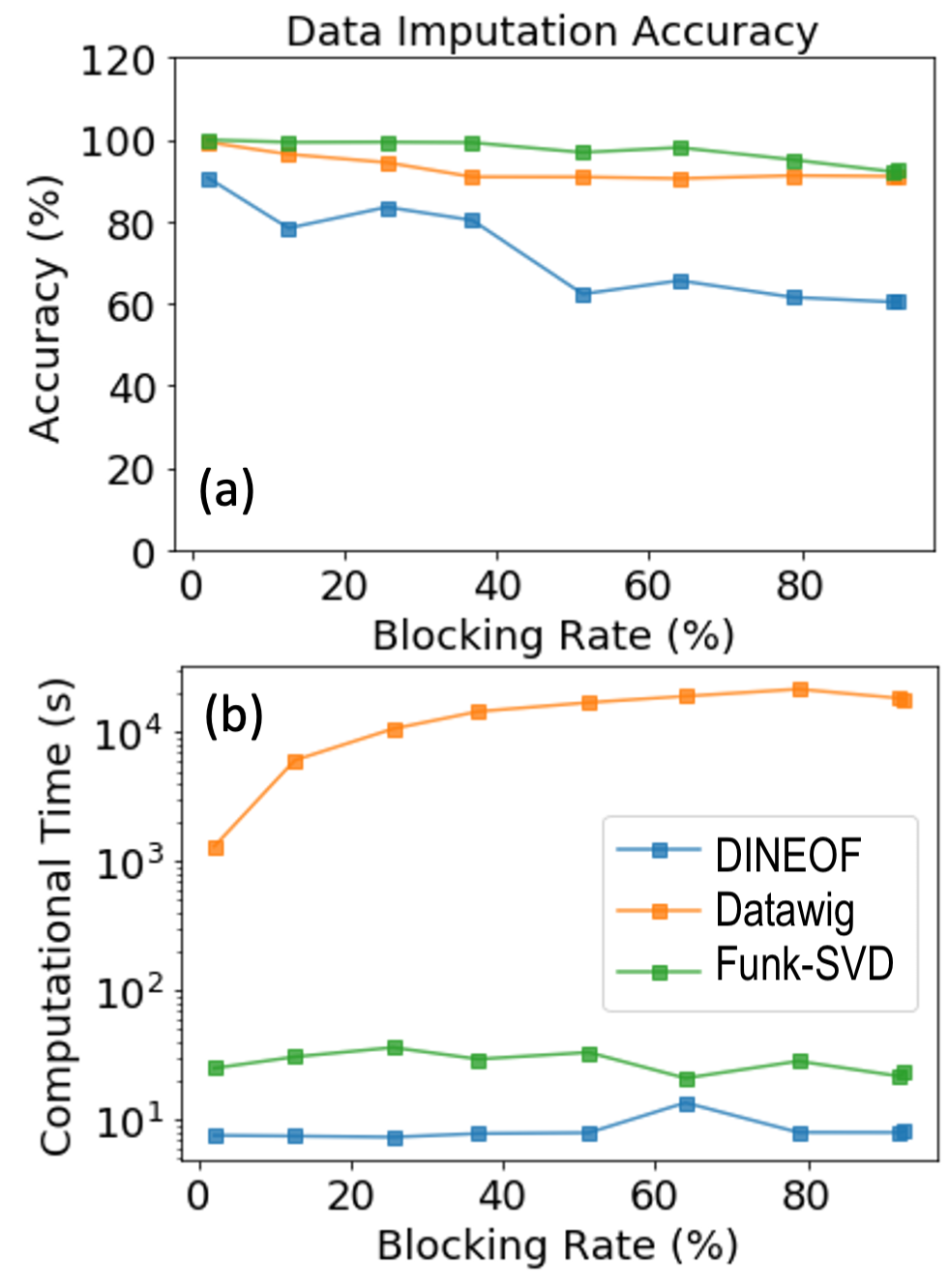}
\caption{The comparison of the data imputation accuracy (a) and computational time (b) for the cloud filling task.}
\label{fig:performance}
\end{figure}

The new method of cloud data imputation is compared with the traditional DINEOF method and the deep learning algorithm – Datawig \cite{biessmann_datawig_2019}. Figure \ref{fig:result} shows that DINEOF was able to recover the missing data at the low blocking rates but the accuracy quickly decreases with the increase of the blocking rate, whereas Datawig was able to recover the data even at the high blocking rates, but the recovered landscape is noisy. The proposed Funk-SVD method showed remarkable performance –- the recovered landscape compares well with the ground truth (the blocking rate of 0) and it remains the coherence of the natural landscape. The F1 scores of the different methods are  compared in Figure \ref{fig:performance}a and the Funk-SVD method is found better than the deep learning method and significantly outperformed the traditional DINEOF method. In addition, Funk-SVD could achieve a high filling accuracy even with high blocking rates. Although the accuracy of the deep learning and Funk-SVD is similar, the Funk-SVD method used much less computational time (Figure \ref{fig:performance}b), which was in the order of 30 seconds -- the deep learning method involves extensive training, requiring four cores of the CPU to run for multiple days. The results show that the proposed Funk-SVD method is able to fill the cloud data with high accuracy and efficiency and has a potential to improve the current practice.

\section*{Discussion}
DINEOF is a widely applied, and probably the most popular, cloud filling algorithm used in remote sensing, but our numerical experiment shows that this benchmark method has difficulty to handle complex and categorical data, which is a common situation in the coastal area. Here, a theoretical analysis is performed to explore the reason behind the under-performance of the method. First, the truncation process can be described as the operation to remain $k$ leading modes in the SVD decomposition,

\begin{multline}
    \mathbf{X}_{M\times N, p}=U_{M\times r,p}\Sigma_{r\times r,p}V_{N\times r,p}^T 
    =U_{M\times k,p}\Sigma_{k\times k,p}V_{N\times k,p}^T+U_{M\times (r-k),p}\Sigma_{(r-k)\times (r-k),p}V_{N\times (r-k),p}^T 
    =\overline{\mathbf{X}}_{M\times N, p}^k+\overline{\mathbf{X}}_{M\times N, p}^{r-k}.\\
\end{multline}

Substituting this expression into Equation \ref{eq:x_bar}, we can obtain the general expression of the iteration,
\begin{equation}
    \mathbf{X}_{M\times N,p+1} 
    =\mathbf{X}_{M\times N,0}+K_{M\times N}\mathbf{X}_{M\times N, p}K_{M\times N}-K_{M\times N}\overline{\mathbf{X}}_{M\times N, p}^{r-k}K_{M\times N}.
\end{equation}
The original matrix can be seen as the ground truth modified by an error matrix $\Delta X_{M\times N}$,
\begin{equation}
    \mathbf{X}_{M\times N,0}=\mathbf{X}_{M\times N}+\Delta \mathbf{X}_{M\times N}
\end{equation}

Combining the iteration scheme of Eqs. \ref{eq:K}-\ref{eq:x_bar}, we can obtain the general formula for the iteration step of $p$,
\begin{equation}\label{eq:prop}
    X_p=X+\Delta X+\sum_{i=1}^{p}K^i(X+\Delta X-\overline X_{p-i}^{r-k})K^i.
\end{equation}
Here, the dimension notation $M\times N$ is omitted for brevity.

This error propagation analysis of Equation \ref{eq:prop} indicates that DINEOF cannot ensure the reduction of the error $\Delta X$ in the iterative operations to converge to the true value, $X$. This analysis is consistent with the observation in Figure \ref{fig:result} that DINOEF could lead to large errors in the final result. The other two methods compared in this study involves minimizing the difference between the result and the original matrix and they can restrain the growth of the error to a limit. Nevertheless, the current framework of the error propagation provides an opportunity and a tool to invent new algorithms to improve mode decomposition based algorithms.

Another potential area to improve the existing algorithms is to explore the use of temporal and spatial information to inform the data imputation. The present numerical experiment re-arranged the image to a vector and stacked the vectors to a matrix. This re-arrangement may lead to the loss of the temporal spatial relationship of the original matrix. Specifically, the horizontal neighboring relationship between pixels was remained but the vertical neighboring relationship was disconnected. This vertical relationship might be still recovered in the latent space for certain methods, but may be lost in deep learning methods which don't consider the position of the pixels. In addition, the revisit frequency of 14 days may not reflect the short processes such as tidal cycles, but it can still capture the long processes such as the seasonal and annual variation of the earth surface's dynamics. Such dynamics information might be harnessed in the emerging mode decomposition methods such as Dynamic Mode Decomposition \cite{schmid_dynamic_2010}, which has been shown effective in coastal process analysis \cite{wang_interactions_2017, li_efficient_2020} and can potentially improve the existing methods.

An important factor that could affect the performance of the algorithms but has not been fully discussed in the paper is the data type. The landscape data prepared by the USGS follows a predetermined spectrum analysis rule. In this rule, different bands of the satellite imaging are used to classify pixels into different landscape types. Although the obtained data is categorical, they have an order arranged following the wetness of the landscape, i.e. water-wetland-land. Strictly speaking, the EOF-based methods are not fully appropriate to be applied to categorical data and hence the comparison with other methods might be unfair, but we found those methods still could produce reasonable results because of the data order -- the categorical data can be treated as continuous and a rounding scheme can be used to convert the continuous numerical data into the orderly categorical data. While the deep learning method doesn't have this difficulty, because the algorithm can be designed to handle categorical data.

\section*{Conclusion}
A numerical experiment was conducted to explore the algorithms to fill the cloud data of satellite-based landscape imaging in a coastal area. Three cloud filling schemes were compared including the benchmark method of DINEOF, a deep learning algorithm of Datawig, and a recommender system-inspired algorithm of Funk-SVD.

We discovered that Funk-SVD achieved the best accuracy with high efficiency of computational  resources: the accuracy is the best among the three compared schemes and the computational speed is similar to DINEOF but much faster than Datawig. The deep learning method of Datawig could reach the similar level of accuracy but need a training time two order longer than Funk-SVD. Funk-SVD is also found capable to recover sparse dataset, which is a common situation found in recommender systems.

A theoretical framework was developed to analyze the traditional DINEOF method and found the method could accumulate unwanted error in the iteration procedure. This theoretical analysis has a potential to inspire the design of the next efficient and accurate data imputation algorithm to improve both cloud filling and recommender system tasks.

\section*{Acknowledgement}
The author acknowledges the funding support by the Rutgers EOAS seeding grant and the US Department of Transportation through the Center of Advanced Infrastructure Technology. The author would like to thank Ms. Katherine Fang for assisting the preparation of the satellite data for the analysis.







\bibliographystyle{elsarticle-num}
\bibliography{references.bib}







\end{document}